\documentclass[conference]{IEEEtran}
\IEEEoverridecommandlockouts
\usepackage{cite}
\usepackage{amsmath,amssymb,amsfonts}
\usepackage{algorithmic}
\usepackage{graphicx}
\usepackage{textcomp}
\usepackage{xcolor}
\usepackage{bm}
\usepackage{authblk}

\usepackage{multirow}
\def\BibTeX{{\rm B\kern-.05em{\sc i\kern-.025em b}\kern-.08em
    T\kern-.1667em\lower.7ex\hbox{E}\kern-.125emX}}
\begin{document}

\title{Structure Regularized Attentive Network for Automatic Femoral Head Necrosis Diagnosis and Localization
\thanks{Lingfeng Li and Huaiwei Cong contribute equally to this work. Corresponding author: Jinpeng Li (lijinpeng@ucas.ac.cn). This work was supported in part by National Natural Science Foundation of China (62106248).}
}

\author[1,2]{\textbf{Lingfeng Li}}
\author[1,2]{\textbf{Huaiwei Cong}}
\author[3]{\textbf{Gangming Zhao}}
\author[4]{\textbf{Junran Peng}}
\author[5]{\textbf{Zheng Zhang}}
\author[1,2,*]{\textbf{Jinpeng Li}}

\affil[1]{HwaMei Hospital, University of Chinese Academy of Sciences (UCAS), Ningbo, China}
\affil[2]{Ningbo Institute of Life and Health Industry, UCAS, Ningbo, China}
\affil[3]{Department of Computer Science, University of Hong Kong, Hong Kong, China}
\affil[4]{National Laboratory of Pattern Recognition, Institute of Automation, Chinese Academy of Sciences, Beijing, China}
\affil[5]{Department of Computer Science and Technology, Harbin Institute of Technology, Shenzhen, China}
\affil[*]{Email: lijinpeng@ucas.ac.cn}

\maketitle

\begin{abstract}
In recent years, several works have adopted the convolutional neural network (CNN) to diagnose the avascular necrosis of the femoral head (AVNFH) based on X-ray images or magnetic resonance imaging (MRI). However, due to the tissue overlap, X-ray images are difficult to provide fine-grained features for early diagnosis. MRI, on the other hand, has a long imaging time, is more expensive, making it 
impractical in mass screening. Computed tomography (CT) shows layer-wise tissues, is faster to image, and is less costly than MRI. However, to our knowledge, there is no work on CT-based automated diagnosis of AVNFH. In this work, we collected and labeled a large-scale dataset for AVNFH ranking. In addition, existing end-to-end CNNs only yields the classification result and are difficult to provide more information for doctors in diagnosis. To address this issue, we propose the structure regularized attentive network (SRANet), which is able to highlight the necrotic regions during classification based on patch attention. SRANet extracts features in chunks of images, obtains weight via the attention mechanism to aggregate the features, and constrains them by a structural regularizer with prior knowledge to improve the generalization. SRANet was evaluated on our AVNFH-CT dataset. Experimental results show that SRANet is superior to CNNs for AVNFH classification, moreover, it can localize lesions and provide more information to assist doctors in diagnosis. Our codes are made public at https://github.com/tomas-lilingfeng/SRANet.
\end{abstract}


\section{Introduction}
Avascular necrosis of the femoral head (AVNFH) is a common clinical disease characterized by progressive pain and disabling degeneration of the hip joint~\cite{yoo2008long}. 
In China, about 100,000 new patients are diagnosed with AVNFH each year, and the total number of AVNFH patients reaches eight million. Given that the direct cost to patients for undergoing primary total hip arthroplasty per hip without revision is between \$8,000 and \$10,000, which suggests that the overall cost of treating all the cases can reach up to \$64 billion. This bring a huge financial burden on the society~\cite{zhao2015prevalence}. The early diagnosis and intervene is the key to reduce the AVNFH damage.

In the clinic, the first imaging modality for AVNFH diagnosis is X-ray due to its low cost, accessibility, and imaging speed. Nevertheless, the overlapping of tissues in the X-ray image makes it difficult to demonstrate occult disease features.
When more imaging evidences are needed, extra exams will be performed, i.e., computerized tomography (CT) and magnetic resonance imaging (MRI). MRI has been shown to be the most sensitive method of detecting the presence of early AVNFH~\cite{shimizu1994prediction}. Compared with CT, MRI is time-consuming and expensive, so that CT has become a feasible choice for the screening of AVNFH. 
Currently, it is very difficult and laborious for experts to seek for the faint CT manifestations of AVNFH, especially at early stages. In order to help doctors diagnose AVNFH faster and more accurately, the computer-assisted diagnosis systems are urgently needed.

With the rapid development of deep learning (DL), deep neural networks (DNN) have demonstrated exceptional capabilities in solving diagnostic problems. Using annotated data for supervised learning, some DNN algorithms have reached or exceeded the performance of human experts in various fields, for example, ophthalmology~\cite{ting2017development}~\cite{li2018efficacy}, dermatology~\cite{esteva2017dermatologist} and orthopedics~\cite{chung2018automated}. 
Although few studies~\cite{li2020deep,chee2019performance,liu2021magnetic} have attempted to perform AVNFH diagnosis using deep learning, none of them have used the CT modality. 
This may be due to the absence of CT datasets with sufficient scale and high quality annotations, making DNN algorithms for 3D classification not accurate for AVNFH diagnosis.
Moreover, previous studies on deep learning based diagnosis of AVNFH have used end-to-end networks for classification, which lack a specific basis for assisting physicians in diagnosing diseases and have low interpretability of results.

In this paper, we propose a novel deep neural network with a gated self-attention mechanism that can weakly supervise the localization of lesions and incorporate mask patch a priori constraints, called  Structure Regularized Attentive Network (SRANet), for classifying femoral regions using CT. 
SRANet uses a neural network-based permutation-invariant aggregation operator corresponding to the attention mechanism.  This allows to explain the contribution of instances to the final decision and can be further used to highlight possible regions of interest. In addition, SRANet use a novel anatomically constrained attention weighting for combining anatomical knowledge with a data-driven permutation-invariant aggregation operator. SRANet is conceptually similar to the attention-based deep multiple instance learning~\cite{ilse2018attention}. The main difference is that we exploit a new method to generate the attention weights.
The main contributions of this work are:
\begin{itemize}
\item We propose the SRANet, which allows for the weakly supervised localization of femoral head necrosis lesions when conducting the AVNFH classification. This makes the diagnosis of AVNFH more intuitive and interpretable to users than end-to-end neural networks.
\item We exploit a new attention weighting method with anatomical constraints by segmenting the femoral head region using conventional image processing techniques. The result of the femoral head segmentation is used to constrain the attention region of the neural network-based permutation-invariant aggregation operator.
\item We construct the first AVNFH-CT dataset with careful annotations by experienced radiologists. Experimental results indicate SRANet is superior to 3D DNNs, demonstrating the capacity of deep SRANet in the automatic diagnosis for AVNFH based on CT. The codes have been made publicly available.

\end{itemize}

\begin{table*}
\begin{center}
\caption{Demographic and avascular necrosis of the femoral head (AVNFH) staging information for all datasets. Note that data in parentheses are percentages.} \label{tab1}

\begin{tabular*}{10cm}{cccc}
\hline
   Classification \&                         & Training Cohort & Validation Cohort  &  Test Cohort \\
                            
  Characteristics                 &(n = 1244)        &  (n=155)               &(n=150)    \\
\hline
Negative for AVNFH         & 628             &78                      &75     \\
person                      & 604         & 76                 &70\\
Mean age(y)                 & 44.6 $\pm$ 18.2      & 41 $\pm$ 16.9              &48.2 $\pm$ 19.5 \\
Women                       & 268(42.7)         & 30(38.5)                 &35(46.7) \\
\hline

Positive for AVNFH         & 616             &77                      &75     \\
person                      & 547         & 68                 &65 \\
Mean age(y)                 & 45.4 $\pm$ 16.9      & 40.6 $\pm$ 16.4              &45.5 $\pm$ 17.1 \\
Women                       & 228(30)         & 30(40)                 &32(42.7) \\
\hline
F1 for AVNFH                 & 31             &5                      &4     \\
person                      & 26         & 4                 &3 \\
Mean age(y)                 & 47.5 $\pm$ 15.4      & 31.5 $\pm$ 10.8         &36.5 $\pm$ 13.5\\
Women                       & 5(16.1)         & 4(80)                 &2(50) \\
\hline
F2 for AVNFH                 & 198             &24                      &26     \\
person                      & 182         & 23                 &23 \\
Mean age(y)                 & 39.8 $\pm$ 13.9      & 37.7$\pm$ 11.3    &40.4$\pm$ 13.8 \\
Women                       & 59(29.8)         & 5(20.8)                 &10(38.5) \\
\hline
F3 for AVNFH                 & 199             &23                      &22     \\
person                      & 191             &23                &21 \\
Mean age(y)                 & 39.7 $\pm$ 15.8      & 41.2 $\pm$ 19              & 46.4 $\pm$ 16.8 \\
Women                       & 69(34.7)         & 8(34.8)                 &9(40.9) \\
\hline
F4 for AVNFH                 & 188             &25                      &23     \\
person                      & 172           & 24                 &21 \\
Mean age(y)                 & 55.7 $\pm$ 15.7 & 44.7 $\pm$ 17.3      &52.2 $\pm$ 18.6 \\
Women                       & 95(50.5)         & 12(50)                 &11(47.8) \\

\hline

\end{tabular*}
\end{center}
\end{table*}

\section{Materials And Methods}

\subsection{Dataset}

CT images of 1,510 AVNFH patients from 2013 to 2020 were extracted from our institution's image archiving and communication system. Of these patients, 1,364 had both CT images and magnetic resonance (MR) images containing the femoral head region. Only a few patients have CT image examination results but no MR images.      
To  create a trustworthy data set for model training, 146 of the 2,687 CT images were excluded due to there was no corresponding MR image. In this way, the CT images submitted to our experts for annotation have the corresponding MR images for reference, which can make the annotation more reliable. 

Our team of experts performing the labeling work consists of three orthopedic surgeons with extensive clinical and surgical experience. When labeling femoral head bounding boxes and femoral head categories, each annotator is responsible for 1/3 of the data without overlap, given the relatively low complexity of these tasks. When staging AVNFH, the annotators labeled all femoral heads individually and recorded a list of typical cases at each stage.

We used the  Association Research Circulation Osseous (ARCO) staging system to stage AVNFH~\cite{meier2014bone,yoon20202019}  with the following criteria: (1) Stage~0: normal; (2) Stage~\uppercase\expandafter{\romannumeral1}: CT examination shows no abnormalities, and MRI examination shows spotty signals; (3) Stage~\uppercase\expandafter{\romannumeral2}: Both examinations Visible sclerosis and local cystic transformation; (4) Stage~\uppercase\expandafter{\romannumeral3}: Femoral head morphology is abnormal, and CT shows collapse and cortical fracture; (5) Stage~\uppercase\expandafter{\romannumeral4}: There is inflammation of bones and joints, cartilage damage, MRI examination shows T1W1 degenerative disease, and CT shows the femoral head is severely deformed. Our expert team staged the bounding box containing the femoral head in the CT according to the evaluation method described above. 

In order to achieve better classification results, we take out the bounding box with the staging information from CT. After screening, we obtained 1,549 CT images of the femoral head region, and each image was annotated with staging. We consider stage 0 as the AVNFH absence class;  stage I, stage II, stage III and stage IV as the AVNFH presence classes. Thus, the classes for AVNFH staging include AVNFH absence and the  AVNFH presence class.

In summary, a total of 1,549 CT images of the femoral head region from 1,364 subjects were included in this study. Of these, 155 CT images of the femoral head region were retained for model validation and another 150 CT images of the femoral head region were retained for model testing. The remaining 1,244 CT images of the femoral head region were used as the training set of the model. After our screening, the positive and negative samples of each dataset are roughly balanced, and the number of samples in each stage of the negative samples is also roughly balanced. Only the number of samples in stage I was relatively small. The characteristics, AVNFH staging information, age range, and sex ratio of all datasets are summarized in Table~\ref{tab1}.

\subsection{Image Pre-processing}
DICOM radiographic image file were loaded onto our system using the Pydicom library (version 2.3.0, Python Software Foundation). The CT image is cropped with the bounding box containing the femoral head region as the boundary. Bounding boxes were manually recorded in advance. The CT images in the cropped bounding box are normalized to a common size and pixel intensity distribution. The images were down-sampled and padded to a final size of $128\times 128\times 128$ pixels. The mean pixel intensity and standard deviation of each image were normalized.

In our SRANet, in order to impose a priori constraint on the attention weight, we need to obtain the corresponding mask samples for the training samples. Read and process samples from the training set using the Opencv-python library (version 4.5.5.64, Python Software Foundation). The bone tissue in the femoral head region of the read training set samples is segmented by a number of traditional image processing techniques including adaptive threshold segmentation, erosion and expansion. We call the segmented femoral head region samples as mask samples and match with the CT images of the femoral head region in the training set. 

In this way, we obtain a mask dataset that matches the training set.
the sample size in the mask dataset is $128\times 128\times 128 $ pixels; and it is a binarized image with two values of 0 and 255. We store the mask samples as npy files (a special binary format of the Numpy library)  in the same format as the training set samples. This makes it more efficient to store the mask samples in advance than to process them when the model reads the training set samples to get the matching  mask samples. 

\begin{figure*}
\begin{center}
\includegraphics[width=0.9\textwidth]{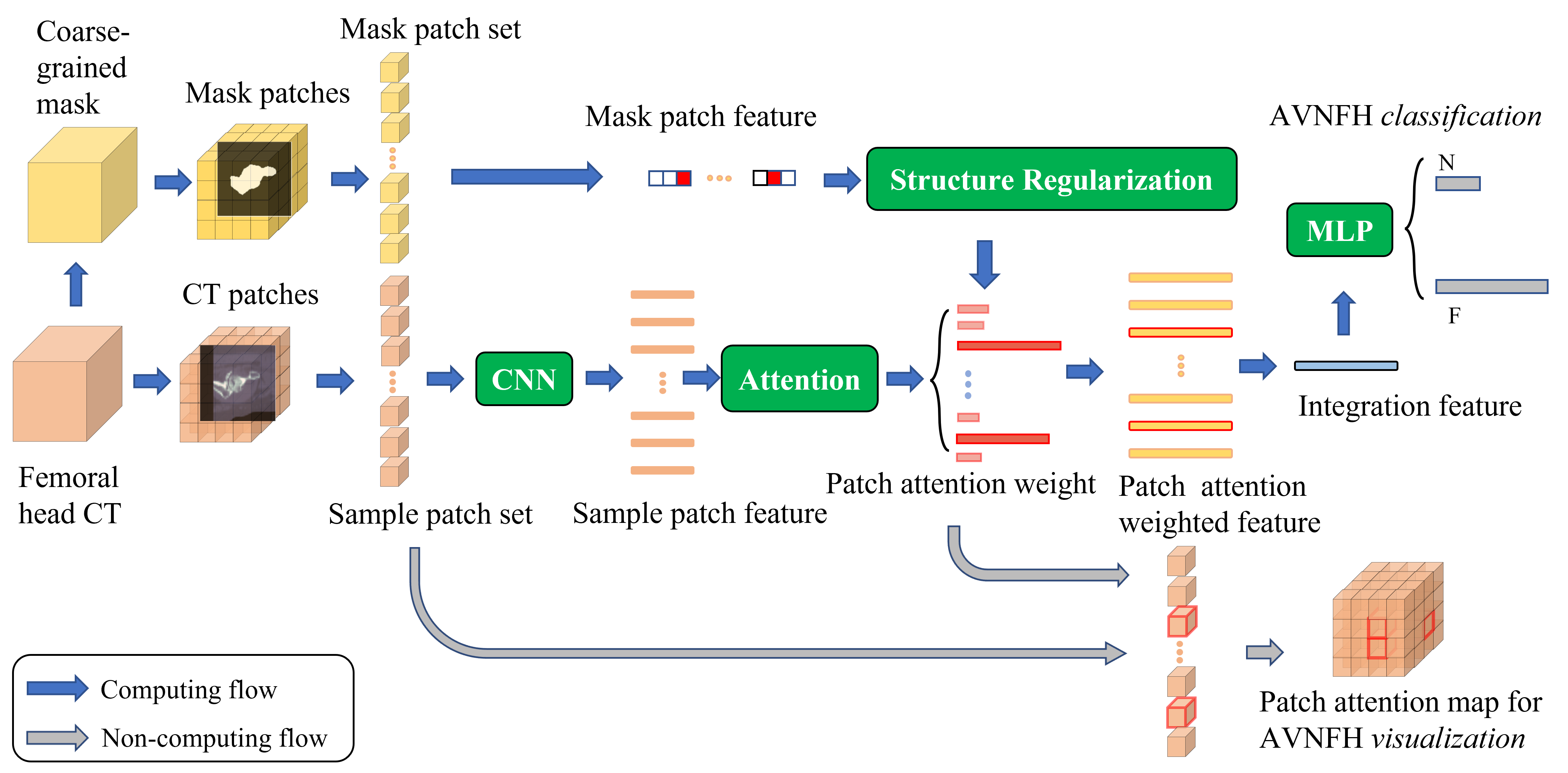}
\caption{Flowchart of the SRANet model. Femoral head CT volumes are processed with conventional image segmentation algorithms to yield coarse-grained segmentation masks (CSM). The original CT, together with the CSM, are divided into mini patches. The mask patch feature is computed and viewed as the structure regularization factor. SRANet utilizes a 3D CNN to capture the features of each CT patch, and then performs a weighted fusion of the patch weights obtained by the attention and structure regularization to achieve more discriminative representations to classify the femoral head region CT images. At the same time, the patch attention maps obtained in this process highlight the high-score regions to assist doctors locate the disease area.} \label{fig01}
\end{center}
\end{figure*}

\begin{table*}   
\begin{center}
	\centering
	\caption{ The performance comparison of classification between our SRANet model and other benchmark models on our test dataset  }\label{tab2}
	\label{tab:1}  
	\begin{tabular}{c|c c c c c}
		\hline\hline\noalign{\smallskip}	
		DNN & Params Size(MB)  & Sensitivity & Specificity & ACC & AUC  \\
		\noalign{\smallskip}\hline\noalign{\smallskip}
		VGG   &   744     & 0.88  &  0.96  & 0.92      &  0.94   \\
		MLP-Mixer & 497     &  0.71    & 0.91    & 0.81           & 0.87     \\
		ViT     &   320     & 0.85   &  0.89      & 0.87         &  0.93  \\
		SRANet  & 26     &  \textbf{0.91}  &  \textbf{0.97}     & \textbf{0.94}      & \textbf{0.95}     \\
		\noalign{\smallskip}\hline
	\end{tabular}
\end{center}
\end{table*}

\begin{figure}
\begin{center}
\includegraphics[scale=0.65]{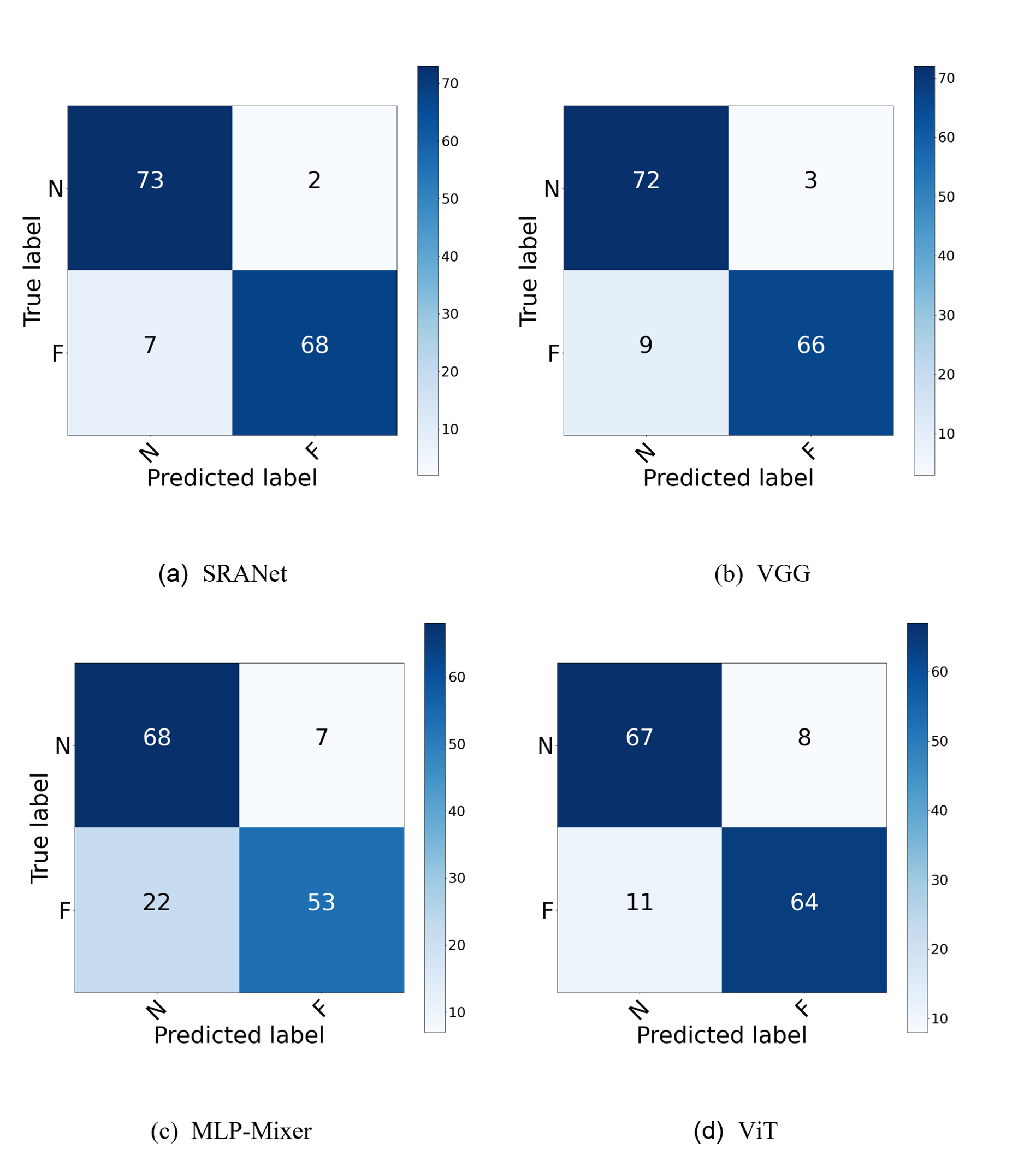}
\caption{Confusion matrices achieved by our SRANet model and other benchmark models on our test dataset.} \label{fig06}
\end{center}
\end{figure}

\begin{figure}
\begin{center}
\includegraphics[scale=0.39]{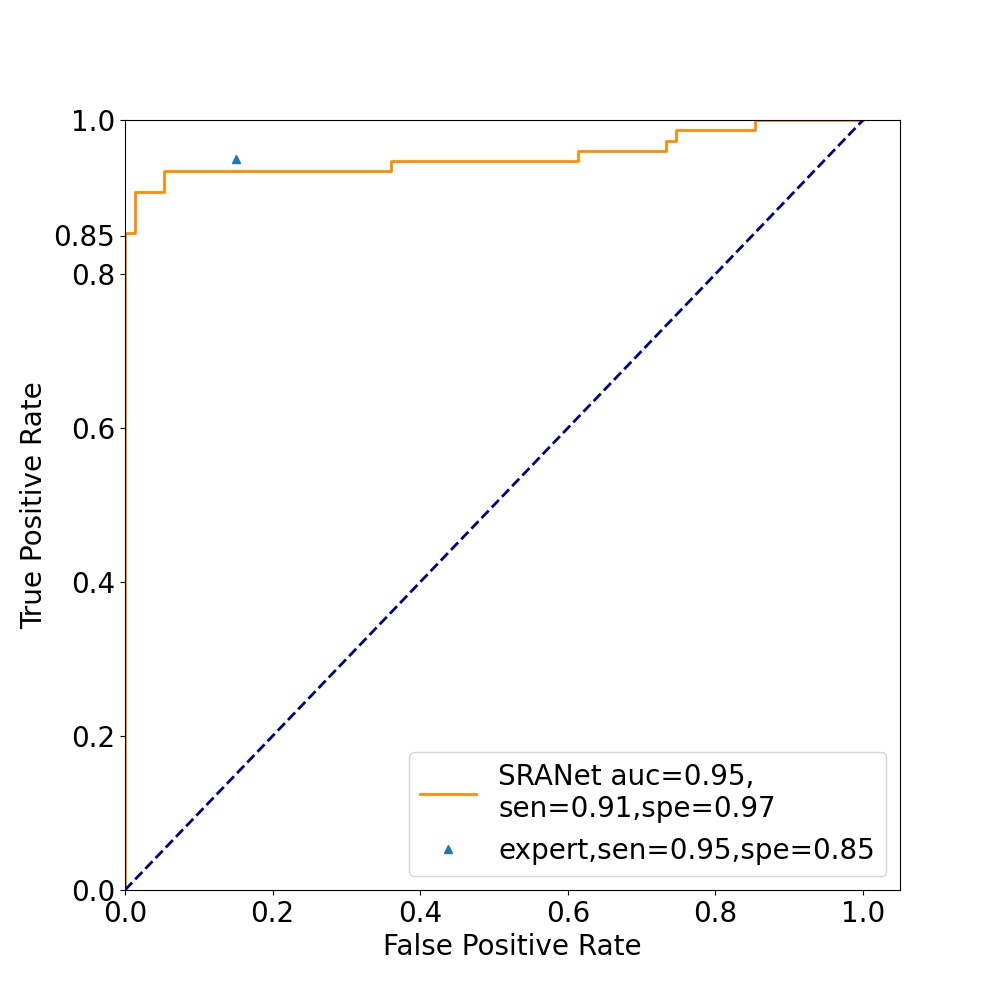}
\caption{The performance comparison of classification between our SRANet model and the expert on our test dataset} \label{fig02}
\end{center}
\end{figure}


\begin{figure}
\begin{center}
\includegraphics[width=0.5\textwidth]{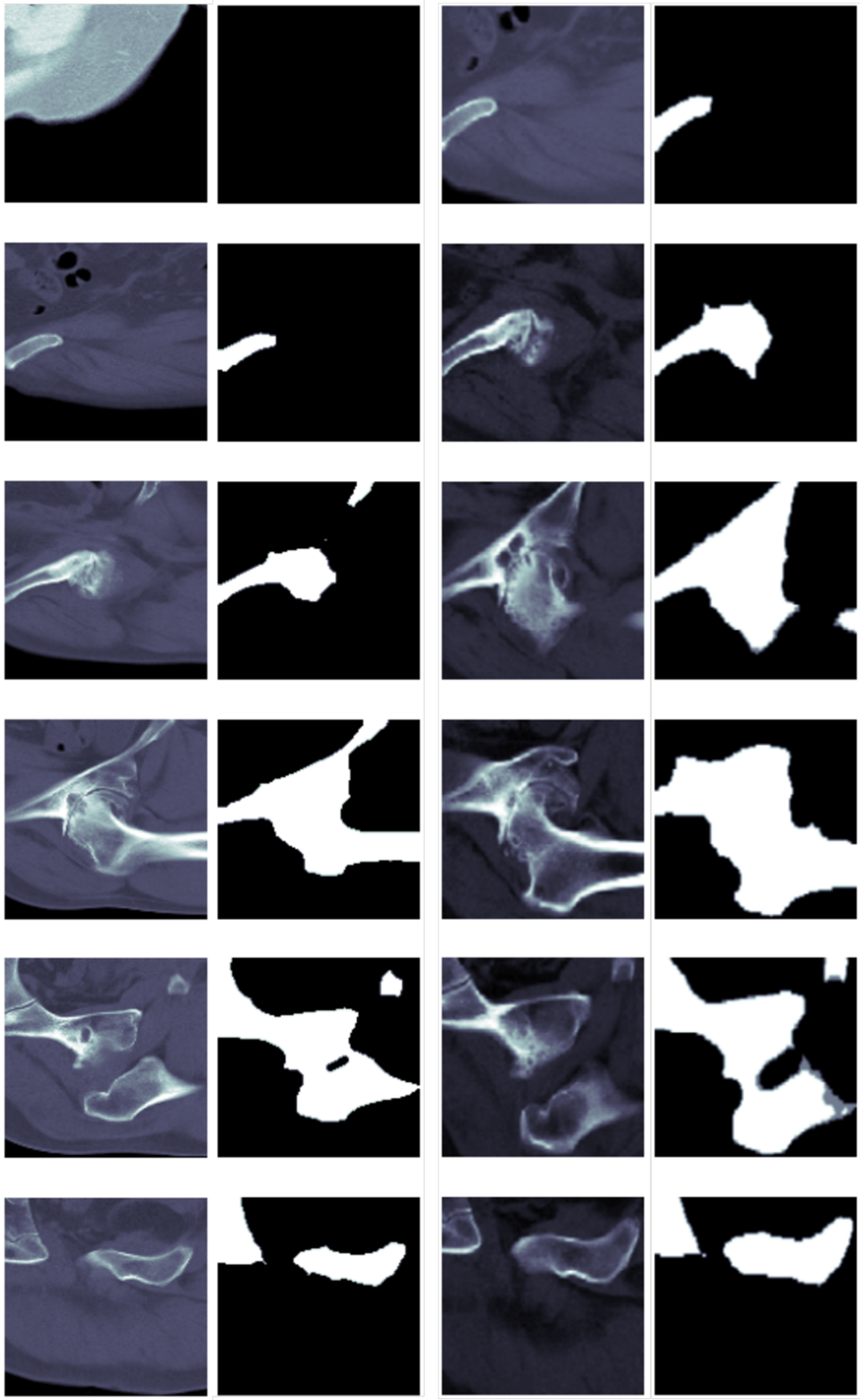}
\caption{Visualization of the data involved in the training. The first column shows the original CT image; the second column shows the mask image of the original CT image after processing; the third and fourth column shows the original CT images and the corresponding mask images after data augment.    }\label{fig03}
\end{center}
\end{figure}

\begin{table}[h]
\begin{center}
\caption{the performance comparison of different CNN feature extractors using the SRANet model and with and without the Structure regularized module. }\label{tab3}
\centering
\begin{tabular}{ c|c | c c |  c   }
\hline
 model & Structure Regularized & Ours & ResNet   &  ACC    \\
\hline
\multirow{4}*{SRANet}  & $\boxtimes$ & $\checkmark$ & $\boxtimes$   &  0.90        \\						
							~	& $\checkmark$ & $\checkmark$ & $\boxtimes$  
								&  \textbf{0.93}   \\
							~	& $\boxtimes$ & $\boxtimes$ & $\checkmark$ 
								& 0.69      \\
							~	& $\checkmark$ & $\boxtimes$ & $\checkmark$ 
								& 0.87   
								\\

\hline
\end{tabular}
\end{center}
\end{table}

\begin{figure*}
\begin{center}
\includegraphics[width=0.9\textwidth]{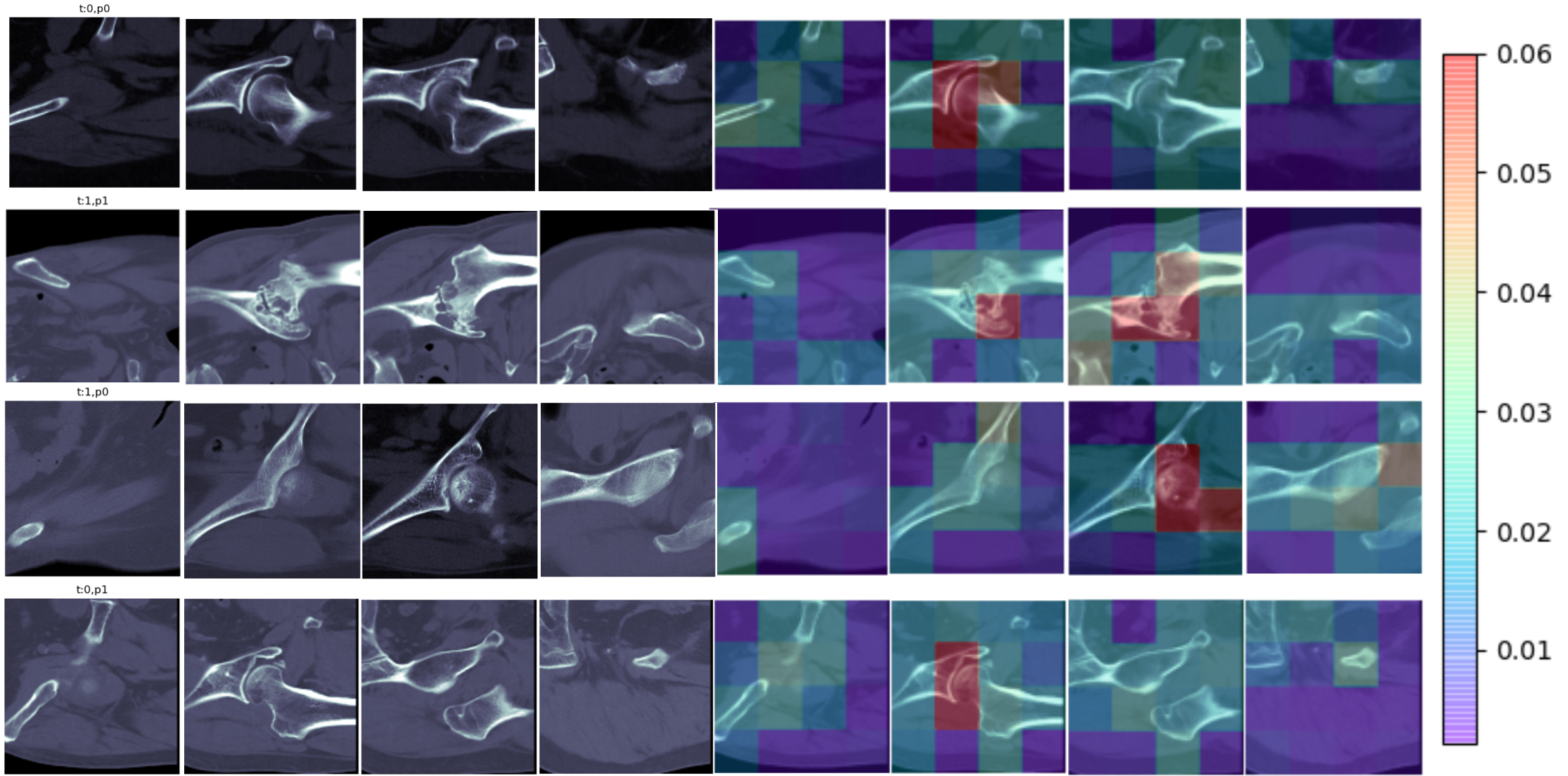}
\caption{Visualization of the patch attention map with structure regularized module. The first  to  fourth rows, respectively, are true negative cases, true positive cases, false negative cases and false positive cases. Each case is a randomly selected sample among that category. The left  half of each row  is the original CT image, and the right half is the corresponding patch attention map. The redder the patch in the patch attention map is, the greater the weight of the feature fusion, and the correspondence between the value of the weight and the color is reflected by the color bar on the right. } \label{fig04}
\end{center}
\end{figure*}

\subsection{SRANet Architecture}
We developed a SRANet model to identify the presence of AVNFH in CT images of the femoral head region. Fig.~\ref{fig01} illustrates the proposed framework for SRANet model.  Firstly, the femoral head region CT images and the matching mask images in the training set are loaded and sequentially chunked into sample patch set and mask patch set, respectively. secondly, CNN streams are used to capture the features of each patch in sample patch set. Then, the patch weights obtained from the attention net and structural regularization are weighted and fused to the sample patch feature to achieve a higher feature presentation to distinguish the CT images of the femoral head region. Finally the downstream neural network further extracts the features and predicts the probability of each class. 
The following will introduce the details of the main modules of SRANet: 1) CNN module; 2) attention module; 3) structure regularized.

1) \textit{CNN Module}: As show in Fig.~\ref{fig01}, the input of our CNN is not a whole CT image, but a Sample patch set of the same size after chunking in order. The input CT images are $ X = \{ x_1, x_2,...,x_n \}, x_i\in \mathbb{R}^{128 \times 128 \times 128} $. Meanwhile, the input image is divided into a patch grid of $P\times P\times P$, and for each grid, the feature vector of the corresponding grid is obtained by CNN. Note that $P$ is an adjustable hyper-parameter ($P=4$ in our experiments). In this way, the entire input  CT image is sliced into sample patch set $X^{\prime}=\{ x_1^{\prime}, x_2^{\prime},...,x_n^{\prime} \},   x_i^{\prime}\in \mathbb{R}^{32 \times 32 \times 32}, n=P*P*P=64 $.  
The image blocks in the sample patch set are extracted with features by our CNN module. In short, the CNN module is computed as:
\begin{equation}
l_1(x_i^{\prime}) =  M(R(f_{1}^{5\times 5\times 5}(x_i^{\prime}))) \label{eq1}    
\end{equation}
\begin{equation}
f_{CNN}(x_i^{\prime}) =  M(R(f_{2}^{5\times 5\times 5}(l_1(x_i^{\prime}))))\label{eq2}
\end{equation}
where R denotes the relu function and $ f^{5\times 5\times 5}$ represents a filter size of $5\times 5\times 5$ for the convolution operation and is a 3D convolution kernel. The M is denoted as max pooling operation, which performs dimension reduction and compression on the input feature map to speed up the computation.

2) \textit{Attention Module}: As shown in Fig.~\ref{fig01}, the Sample patch set obtained the sample patch feature $H=\{h_1,h_2,...,h_n\}, h_i \in \mathbb{R}^M$ after extracting them by CNN module.  We  use a weighted average of instances (low-dimensional embeddings) to fuse the obtained feature vectors of each patch where weights are determined by a neural network. Additionally, the weights must sum to 1 to be invariant to the size of instances.

\begin{equation}
    Z=\sum_{i=1}^N a_i h_i \label{eq6}
\end{equation}
where
\begin{equation}
f_k=tanh(K h_i^{\top}) \label{eq3}
\end{equation}
\begin{equation}
f_q=sigm(Qh_i^{\top}) \label{eq4}
\end{equation}
\begin{equation}
    g_i=w^{\top}(f_k \odot f_q  )
\end{equation}
\begin{equation}
a_i=\frac{g_i-min(g)}{\sum_{i=1}^N (g_i-min(g)) }\label{eq5}
\end{equation}
where $g=\{g_1,...,g_n\}, g_i\in \mathbb{R} $, $w\in \mathbb{R}^L$, $K, Q\in \mathbb{R}^{L\times M} $ are parameters, $\odot$ is an element-wise  multiplication and $sigm(\cdot)$ is the sigmoid element-wise non-linearity. The  $tanh(\cdot)$ is the tanh element-wise non-linearity. The  $min(\cdot)$ is the minimum value in the parentheses. $Z \in \mathbb{R}^{M}$ is the integration feature and $a_i h_i \in \mathbb{R}^{M} $ is the patch attention weighted feature. $a_i \in \mathbb{R}$ denotes the patch attention weight in Fig.~\ref{fig01}.

3) \textit{Structure Regularization}:
As introduced in Section II, the stage of AVNFH can be divided into five phases from Stage 0 to Stage~\uppercase\expandafter{\romannumeral4}. The stages greater than Stage~I can be examined  Visible sclerosis, local cystic transformation,collapse and deformed.According to the clinical experience, human physicians generally diagnose AVNFH by examining the femoral head in CT images. It is intuitive to consider that these abnormalities are focused on bone tissue, so it is important to include the corresponding characteristics of the bone tissue in the patch.  Therefore, we can calculate whether the patches in the Sample patch set contain bone tissue as a priori evidence for the weighting factor of the corresponding feature fusion.
\begin{equation}
    f_m=\{f_1,...,f_n\},     f_i=\langle h_i , m_i \rangle
\end{equation}
where the Mask patch set is $M=\{ m_1,m_2,...,m_n \},m_i\in \mathbb{R}^{32 \times 32 \times 32}, n=P*P*P=64 $, denoted as the mask matrix corresponding to the bone tissue in the Sample Patch set.$\langle \cdot,\cdot \rangle$  calculate whether  there is overlap, if it exists count as $1$ otherwise $0$.

\subsection{Loss Function}
It is noted that there are two supervisions that perform on the SRANet. One supervision is to force SRANet to learn the presence of AVNFH by fused features after weighted averaging. the corresponding loss function is the cross entropy loss function, defined as follows:
\begin{equation}
    l_c= y \cdot log(p) +(1-y) \cdot log(1-p)
\end{equation}
where $y$ denotes the label of the sample, the positive class is 1 , the negative class is 0. $p$ denotes the probability that the sample is predicted to be a positive class, which is obtained from the integration feature $Z$ predicted by the classification neural network.
Another supervision is that the structural regularization module learns a priori knowledge of the bone tissue structure of the CT images of the femoral head region. The Coarsc-grained mask obtained by traditional image processing techniques is used to constrain the weighted average coefficients of the fused features.
We define the loss function $l_s$ as follow:
\begin{equation}
    l_s=\{l_1,...,l_n\}
\end{equation}
\begin{equation}
    l_i= -[f_i \cdot log(\sigma (g_i)) + (1-f_i) \cdot log(1-\sigma (g_i))]
\end{equation}
the final loss function of SRANet can be defined as follow
\begin{equation}
    loss= \alpha l_c + (1-\alpha) l_s 
\end{equation}
where, $\alpha$  denotes the weight, and $\alpha \in (0, 1)$.

\section{Experiments and Results}

\subsection{Experimental Setting}
The model was trained on our femoral CT training dataset  using the adamW optimizer. Within a total of 200 epochs, the learning rate starts from 0.0001 and decreases by 10 times after every 50 epochs. The weight decay and the momentum are 0.0001 and 0.9, respectively. Training and evaluation of the model were performed on a server running a 64-bit Linux operating system (Ubuntu 16.04) with an intel Xeon E5-2680 CPU with 256 GB DDR4 RAM and four Nvidia Tesla V100 graphic cards (Nvidia driver 440.33.01) with 5120 CUDA cores and 32 GB HBM2 RAM. The proposed method and all the comparing methods are implemented with PyTorch~\cite{paszke2017automatic}.

\subsection{Evaluation Metrics}
In this study, a dichotomous task, normal/abnormal classification, was performed on CT images of the femoral head region, and the experimental results of the dichotomous task are reported. For the binary classification tasks, the data are relatively balanced, of which 80\% for each class is used for training,
10\% for validation and the rest 10\% for testing. We adopt sensitivity (Sen), specificity (Spe), accuracy (Acc)~\cite{wilson1927probable}, and area under the curve (AUC)~\cite{bradley1997use} to evaluate the performance. In addition, for the four-class respiratory sound classification, the official score that is the average of sensitivity and specificity is used as the evaluation metric. 
\begin{equation}
    Sen=  \frac{TP}{TP+FP}
\end{equation}
\begin{equation}
    Spe=  \frac{TN}{TN+FP}
\end{equation}
\begin{equation}
    Acc=  \frac{TP+TN}{TP+TN+FP+FN}
\end{equation}
where $TP$, $TN$, $FP$ and $FN$ are true positive, true negative, false positive and false negative, respectively.

\subsection{Comparison with the Reference Model}
CNN-based methods have been proved very effective in image classification, such as VGGNet~\cite{simonyan2014very} , Mlp-mixer~\cite{tolstikhin2021mlp} and ViT~\cite{dosovitskiy2020image}, showing promise in medical imaging diagnosis. 
Table~\ref{tab2} shows the comparison of our model with the benchmark models VGGNet, Mlp-mixer and ViT in normal/abnormal classification. It is worth noting that since the input is a whole CT image, we have made some changes to the benchmark model to fit the input data. The results confirm the superiority of our proposed model over the benchmark models. Besides, confusion matrix results are also provided in Fig.~\ref{fig06}.

\subsection{Comparison with the expert}
Fig.~\ref{fig02} shows the  receiver  operating  characteristic (ROC) curve describing the diagnostic performance of the  SRANet model for  determining the presence or absence of  AVNFH~\cite{hanley1983method} . The ACC for the SRANet model was 0.94, the specificity is 0.97 and the sensitivity is 0.91.  For comparison, point estimates representing the sensitivity and specificity of  clinical orthopedic surgeon with many years of clinical experience  to determine the presence or absence of AVNFH are plotted on the graph and located below the ROC curve of  the SRANet model. The ACC of the expert was 0.90, the specificity was 0.85 and the sensitivity was 0.95. It can be seen that the clinician orthopedic surgeon's estimate of sensitivity was stronger than the model, but the accuracy and specificity were significantly lower than the model, indicating the ability of the machine to assist the clinical radiologist in screening for femoral head necrosis.

\subsection{Visualization of Patch Attention Map}
There are some previous studies related to the diagnosis of AVNFH, and the data sets in these studies are both large and small, using X-ray and MR images. What they all have in common is the use of end-to-end neural networks, which allows the results predicted by the network to be used only as a reference by experts. However, our SRANet model can show the basis of model judgment by visualizing the patch attention map, which can assist the expert to speed up the diagnostic process and improve the rationality of the diagnostic process by focusing on the highlighted parts of the input CT images of the femoral head region. As shown in Fig.~\ref{fig04}, the patch attention map in our SRANet model has a higher interpretability compared with other 
end-to-end benchmark model.

\subsection{Ablation Study}
We investigate the effectiveness of the Structure Regularized module. As shown in Table~\ref{tab3}, on the test dataset, having the Structure Regularized module outperforms not containing it. Therefore, the Structure Regularized module is used in our experiments. We also investigated whether the CNN module should use the feature extraction as shown in Section II or use the deeper 3D ResNet module. As shown in Table~\ref{tab3}, the CNN module we currently use works better.

\section{Conclusion}
 In this paper, we attempt to developing a 3D deep learning model to diagnose AVNFH based on CT. To achieve this, we constructed the first CT dataset of AVNFH staging and designed the SRANet model for the diagnosis of femoral head necrosis as well as a more interpretable visual patch attention map to assist physicians in the diagnosis. Comparative and ablation experiments showed that our proposed model was not only smaller than other models in terms of model size, but also outperformed other benchmark models and experts' assessments. The visualization of patch attention map in our SRANet model has the potential to assist physicians in diagnosis in a more intuitive manner.


{
\bibliographystyle{IEEEtranS}
\bibliography{egbib}
}
\end{document}